\documentclass{article} 

\usepackage{preprint,times}
\iclrfinalcopy

\usepackage{amsmath,amsfonts,bm}

\def\eqref#1{equation~\ref{#1}}

\def\1{\bm{1}}

\DeclareMathAlphabet{\mathsfit}{\encodingdefault}{\sfdefault}{m}{sl}
\SetMathAlphabet{\mathsfit}{bold}{\encodingdefault}{\sfdefault}{bx}{n}

\usepackage{hyperref}
\usepackage{url}

\usepackage[utf8]{inputenc}
\usepackage[T1]{fontenc}
\usepackage{algorithm}
\usepackage{algorithmic}
\usepackage{graphicx}
\usepackage{booktabs}
\usepackage{multirow}
\usepackage{amsmath}
\usepackage{amsfonts}
\usepackage{enumitem}
\usepackage{url} 
\usepackage{xspace}
\usepackage[table]{xcolor}
\usepackage{hyperref}
\usepackage{epigraph}
\usepackage{amsthm}
\usepackage{amssymb}
\usepackage{arydshln}
\usepackage{bbm}
\usepackage{wrapfig}
\usepackage[most]{tcolorbox}

\definecolor{langlightblue}{rgb}{0.3, 0.65, 1}
\definecolor{langblue}{rgb}{0, 0.4, 0.8}
\definecolor{langmildblue}{rgb}{0.0, 0.45, 0.73}
\definecolor{langdarkblue}{rgb}{0.0, 0.0, 0.61}
\definecolor{langred}{rgb}{0.81, 0.09, 0.13}
\definecolor{langlightgreen}{rgb}{0.80, 0.97, 0.85}
\definecolor{langgreen}{rgb}{0.18, 0.55, 0.34}
\definecolor{langdarkgreen}{rgb}{0.0, 0.45, 0.38}
\definecolor{bingpink}{rgb}{1.0, 0.41, 0.71}
\hypersetup{
    colorlinks=true,
    linkcolor=langdarkgreen,
    citecolor=langgreen,
    urlcolor=langdarkgreen,
}
\newcommand{\gain}[1]{\textbf{\textcolor{langdarkgreen}{#1}}}
\newcommand{\loss}[1]{\textbf{\textcolor{langred}{#1}}}

\makeatletter
\AtBeginDocument{%
  \let\oldref\ref
  \renewcommand{\ref}[1]{%
    \hyperref[{#1}]{\underline{\oldref{#1}}}%
  }%
}
\makeatother

\usepackage{titletoc}
\newcommand\DoToC{%
  \startcontents
  \printcontents{}{1}{\textbf{\large Contents of Appendix}\vskip3pt\hrule\vskip5pt}
  \vskip3pt\hrule\vskip5pt
}

\newtcblisting{promptbox}[2][]{
    enhanced,
    width=\linewidth,
    colback=gray!5,
    colframe=gray,
    title=\textbf{#2},
    listing only,
    listing options={
        basicstyle=\normalfont\itshape\small,
        breaklines=true,
        breakatwhitespace=true,
        columns=fullflexible
    },
    #1
}

\title{MedRouter: Demystifying Knowledge Differences Across Medical LLMs for Routing-Based Reasoning}

\author{
  Lang Cao$^1$ \quad Binghang Lu$^2$ \quad Yuhao Shen$^3$ \quad Yue Guo$^1$\\
  $^1$University of Illinois Urbana-Champaign
  $^2$Purdue University \\
  $^3$The Chinese University of Hong Kong, Shenzhen
}

\begin{document}

\maketitle

\begin{abstract}
Medical question answering spans diverse specialties and modalities, and individual medical large language models (LLMs) exhibit distinct strengths across tasks and domains. This heterogeneity suggests that combining specialists may enable broader coverage of medical questions than relying on any single model. However, existing LLM routing methods primarily seek to balance answer quality and inference cost, leaving open how to exploit differences in specialist competence to improve medical reasoning. In this paper, we introduce \textit{MedRouter}, an agentic system that uses an embedding-based multi-label router to select and query specialist LLMs, then passes their responses to a generator to produce the final answer. We further propose \textit{SCALE} (\textit{Specialist Competence-Aware Learning}), a two-stage training framework that first trains the Router with specialist correctness supervision and then optimizes its selections through reinforcement learning. The second stage uses a Performance Gain Reward (PGR) that measures how specialist information affects the generator's answer correctness relative to answering without that information. Experiments on eight text-based and multimodal medical QA benchmarks show that MedRouter outperforms the strongest routing baseline by 8\% in average accuracy. Our analysis of specialist outputs further reveals distinct strengths and complementary question-level coverage, motivating learned routing to combine these capabilities for more comprehensive medical reasoning.
\end{abstract}


\section{Introduction}
Medical tasks span clinical specialties and involve information presented as text, images, or both. To support these tasks, a growing range of medical large language models (LLMs) have been developed \citep{labrak2024biomistral,sellergren2025medgemma,wang2023huatuo}. These models differ in the sources, domains, coverage, and modalities of their training data, the data available to their developers, and the tasks they are designed to perform.
Such differences may shape the knowledge that models acquire and retain \citep{chang2024large}. Therefore, a model may excel in one specialty while lacking knowledge that another model possesses \citep{xie2025medical}. On questions that require this missing information, the other model may answer correctly even if its overall accuracy is lower. Choosing a single model based on average performance could then overlook useful capabilities available in other models. We therefore investigate whether existing medical LLMs have complementary question-answering capabilities and how their responses can be selected and combined to improve medical question answering.

\begin{figure*}[t!]
    \centering
    \includegraphics[width=\textwidth]{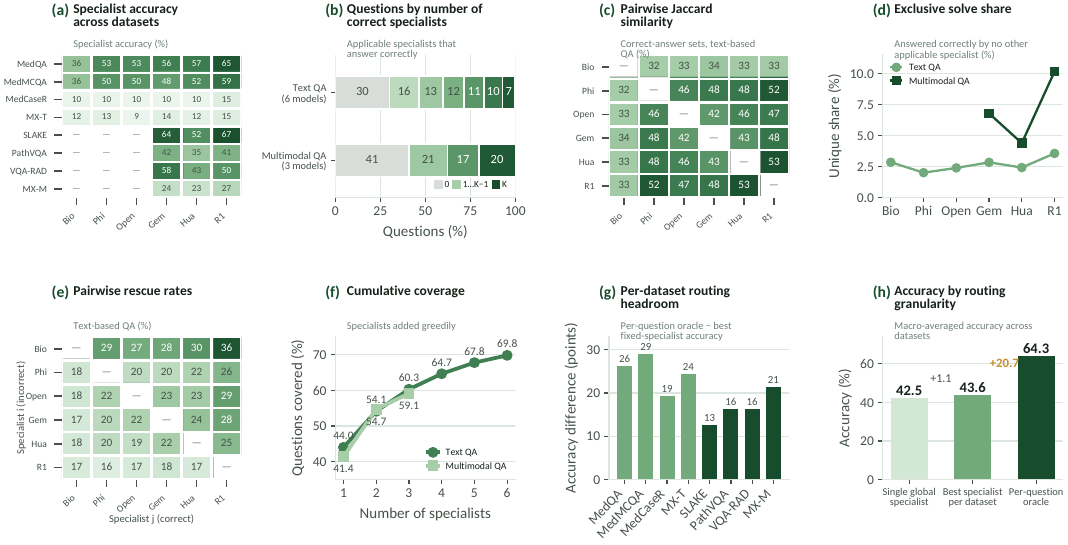}
    \caption{\textbf{Question-level complementarity among medical specialist LLMs.}
(a) Specialist accuracy across datasets.
(b) Distribution of questions by the number of applicable specialists that answer correctly.
(c) Pairwise Jaccard similarity between specialists' correct-answer sets on text-based QA.
(d) Exclusive solve share: questions answered correctly by one specialist but by no other applicable specialist.
(e) Pairwise rescue rates on text-based QA: the fraction of questions answered incorrectly by specialist \(i\) that specialist \(j\) answers correctly.
(f) Cumulative coverage as specialists are greedily added, where a question is covered if at least one selected specialist answers correctly.
(g) Per-dataset routing headroom, defined as per-question oracle accuracy minus the best fixed-specialist accuracy.
(h) Macro-averaged accuracy of a single global specialist, the best specialist per dataset, and a per-question oracle.
The oracle counts a question as correct whenever at least one applicable specialist answers correctly.}
    \label{fig:analysis}
\end{figure*}

To examine this possibility, we analyze specialist responses at the question level, comparing which questions six medical specialist LLMs answer correctly across four text-based QA benchmarks (MedQA, MedMCQA, MedCaseR, and MX-T) and four multimodal QA benchmarks (SLAKE, PathVQA, VQA-RAD, and MX-M). The multimodal evaluation includes only the three vision-capable specialists (see \S\ref{sec:experiments}). Although MediX-R1 \citep{mullappilly2026medixr1} achieves the highest accuracy on all four text benchmarks (Figure~\ref{fig:analysis}(a)), each specialist correctly answers some questions missed by all other models in the applicable pool (Figure~\ref{fig:analysis}(b-g)). To quantify how much additional coverage these differences offer, we compare specialist selection at the dataset and question levels. Selecting the best specialist for each dataset yields 43.6\% micro-averaged accuracy across the eight benchmarks, only slightly above the 42.5\% achieved with one best global specialist. By contrast, a per-question oracle that selects a correct specialist response whenever one is available reaches 64.3\% (Figure~\ref{fig:analysis}(h)). This 20.7-percentage-point gap reflects questions missed by the best fixed specialist for each dataset but answered correctly by another specialist. These findings motivate question-level routing to leverage complementary specialist capabilities for more comprehensive medical reasoning, rather than relying on dataset-level rankings.

Existing routing systems learn to select models at the question level, often to balance answer quality and inference cost \citep{ding2024hybrid,chen2024routerdc,ong2024routellm}. Some approaches also incorporate responses from selected models into answer generation \citep{zhang2025router,qian2025xrouter,xia2025mmedagent}. Building on this line of work, we place greater emphasis on leveraging complementary specialist knowledge for medical reasoning. Moreover, many agentic routing systems use a single LLM for both model selection and answer generation, making it difficult to optimize routing to fully exploit the information provided by specialists.

In this paper, we introduce \textbf{\textit{MedRouter}}, an agentic routing system which separates specialist selection from answer generation. An embedding-based multi-label router selects one or more medical specialist LLMs for each question, and a generator uses their responses together with the original question to produce the final answer. We train the router with \textbf{\textit{SCALE}} (\textit{\textbf{S}pecialist \textbf{C}ompetence-\textbf{A}ware \textbf{LE}arning}), a two-stage procedure. Stage~I uses specialist correctness labels to learn which specialists can answer each question independently. Stage~II then evaluates the selected responses through the generator and updates the routing policy with reinforcement learning. Its Performance Gain Reward (PGR) is computed from the generator's answer correctness with and without the selected responses. Throughout both stages, only the router is updated; the embedding encoder, specialists, and answer generator remain frozen.

We evaluate \textit{MedRouter} using \textit{Qwen3.5-2B} and \textit{Gemma-4-E2B} as answer generators across eight benchmarks. \textit{MedRouter} achieves the highest mean accuracy across benchmarks among the compared methods with both generators, reaching 45.9\% and 46.1\%, respectively. Both exceed the 43.6\% benchmark average of Best-in-Pool, which selects the strongest standalone specialist separately for each dataset. Ablations with \textit{Qwen3.5-2B} show that SCALE outperforms training with either stage alone, supporting specialist correctness supervision followed by generator-level feedback. Further analyses strengthen the motivation for \textit{MedRouter} and demonstrate its advantages.
To sum up, our contributions are:
\begin{itemize}[noitemsep, topsep=1pt, leftmargin=10pt]
    \item We characterize question-level complementarity among medical specialist LLMs and quantify the additional correct-answer coverage available beyond selecting the best specialist for each dataset.
    \item We introduce \textit{MedRouter} for multi-label specialist selection and response integration, and \textit{SCALE}, which trains the router through specialist correctness supervision followed by reinforcement learning from generator-level feedback.
    \item We evaluate \textit{MedRouter} across eight text-based and multimodal medical QA benchmarks, achieving the highest mean accuracy among the compared methods with both answer generators. Out-of-distribution tests and ablations assess generalization and the contribution of each training stage.
\end{itemize}

\section{Related Work}
\noindent \textbf{Reinforcement Learning for Agentic System.}
Reinforcement learning (RL) \citep{Kaelbling1996ReinforcementLA} has recently been widely applied to LLMs to improve reasoning capabilities and agent adaptation \citep{Ouyang2022TrainingLM, guo2025deepseek, liu2025superrl, liu2025bingo}. Representative RL algorithms include Proximal Policy Optimization (PPO) \citep{Schulman2017ProximalPO}, Group Relative Policy Optimization (GRPO) \citep{shao2024deepseekmathpushinglimitsmathematical}, and Generalized Diffusion Policy Optimization (GDPO) \citep{liu2026gdpo}. In terms of agentic systems, DeepRetrieval \citep{jiang2025deepretrievalhackingrealsearch}, Search-R1 \citep{jin2025search}, and S3 \citep{jiang2025s3} apply RL to train models to reason about interactions with search engines for information retrieval and question answering. Meanwhile, Formula-R1 \citep{cao2025fortune} uses RL to enable LLMs to perform symbolic table reasoning by deriving executable spreadsheet formulas,  while DeepCoder \citep{luo2025deepcoder} uses RL for code reasoning and generation tasks.
\\

\noindent \textbf{Router for LLM Selection.}
A key motivation for LLM selection is to balance response quality and inference efficiency, since more capable models can be costly to run. A substantial body of work has explored this trade-off. HybridLLM \citep{ding2024hybrid} and C2MAB-V \citep{dai2024cost} use budget-aware routing and cost-sensitive bandits, respectively. TO-Router \citep{stripelis2024tensoropera} and FORC \citep{vsakota2024fly} predict query-specific model suitability, while FrugalGPT \citep{chen2023frugalgpt}, GraphRouter \citep{feng2024graphrouter}, RouteLLM \citep{ong2024routellm}, and RouterDC \citep{chen2024routerdc} explore cascades, graph-based selection, preference-based routing, and contrastive learning. More recently, Router-R1 \citep{zhang2025router} and xRouter \citep{qian2025xrouter} optimize routing policies with RL. However, our motivation differs: we aim to understand knowledge differences across medical LLMs and develop a system that combines their complementary capabilities to provide more comprehensive medical reasoning.
\\

\noindent \textbf{Medical Reasoning with LLMs.}
LLMs are increasingly being applied to a broad range of medical reasoning tasks, including diagnosis \citep{zhou2025large, yang2024drhouse, cao2026guideskill}, clinical prediction from electronic health records \citep{cao2026ehr}, and clinical evidence synthesis \citep{wang2024trialmind,wang2025leads}. Representative benchmarks include MedQA \citep{jin2020disease} and PubMedQA \citep{jin2019pubmedqa} for medical question answering, as well as PathVQA \citep{he2020pathvqa} and PMC-VQA \citep{zhang2023pmc} for multimodal medical reasoning. To improve medical reasoning capabilities, some studies train LLMs on large-scale medical corpora from diverse sources \citep{sellergren2025medgemma, labrak2024biomistral, wang2023huatuo, shen2026medguidex}. Other studies seek to combine the strengths of multiple models through ensemble learning or multi-agent collaboration. For example, Yang et al. \citep{yang2025large} propose an ensemble framework that aggregates predictions from multiple LLMs through voting. CURE \citep{elshaer2025cure} introduces a confidence-driven ensemble that detects uncertainty in a primary LLM and dynamically routes queries to helper models for collaborative reasoning. Another line of work explores multi-agent collaboration, in which LLMs iteratively refine their predictions through self-consistency and collaboration \citep{shang2025collaboration}. MMedAgent-RL \citep{xia2025mmedagent} uses reinforcement learning to coordinate triage and specialist agents for multimodal medical reasoning. Despite this progress, existing approaches typically rely on a single LLM, combine models through fixed or heuristic strategies, or require costly multi-agent interactions. They do not explicitly learn to exploit the complementary medical knowledge of existing LLMs or to integrate this knowledge effectively for medical reasoning.



\begin{figure*}[t!]
    \centering
    \includegraphics[width=\textwidth]{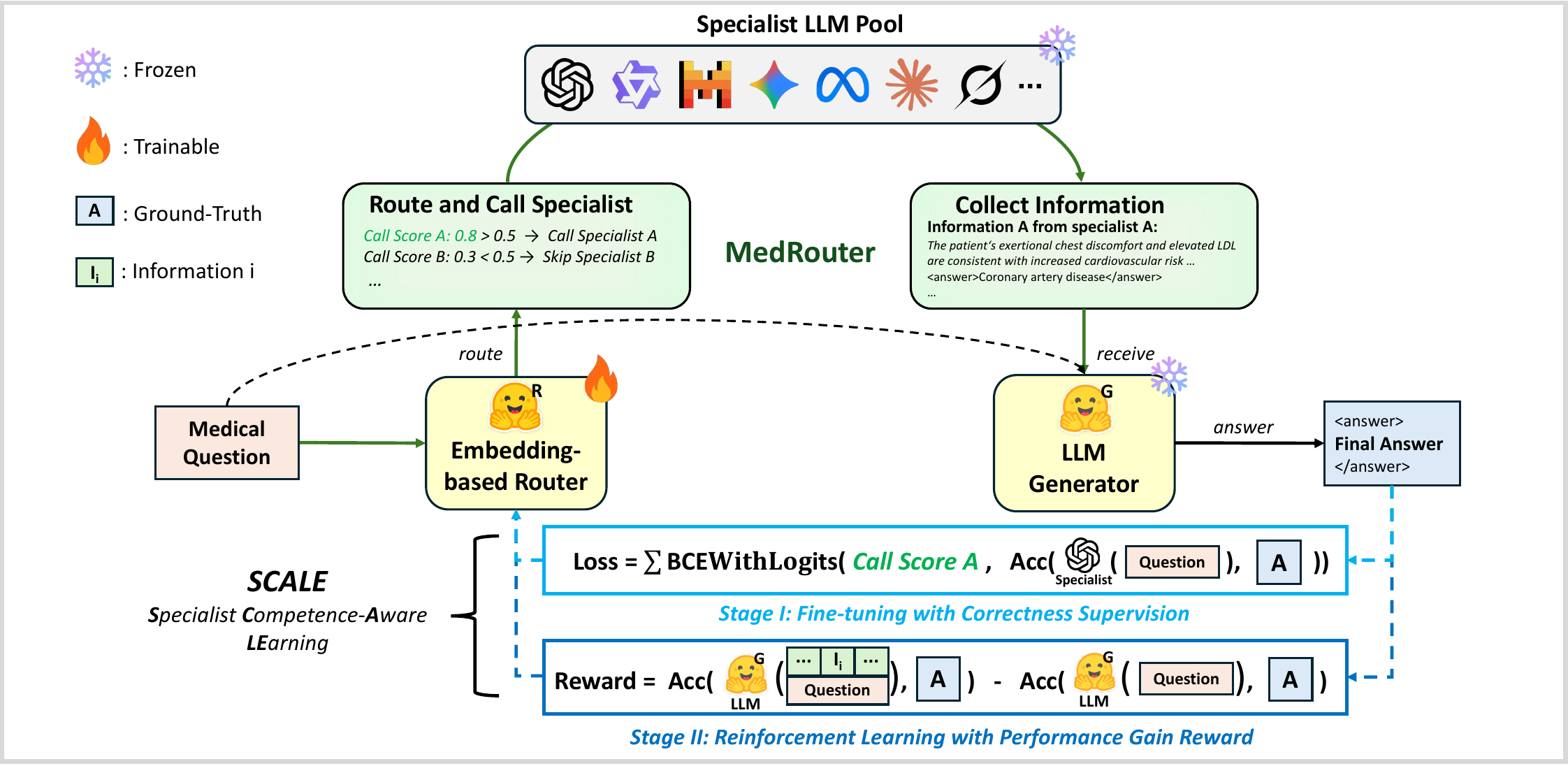}
    \caption{Overview of the \textit{MedRouter} framework and \textit{SCALE} training. Given a medical question, an embedding-based Router predicts call scores and selects specialists from a frozen LLM pool. The frozen LLM generator receives the selected specialists' information and produces the final answer. During training, only the Router is updated: it is first trained with specialist correctness supervision using a binary cross-entropy loss, and then optimized with reinforcement learning using a Performance Gain Reward that compares answer correctness with and without specialist information.}
    \label{fig:main}
\end{figure*}

\section{Methodology}

We introduce \textit{MedRouter}, an agentic system that routes medical questions to a pool of specialist LLMs and aggregates their responses to support medical reasoning, as illustrated in Figure~\ref{fig:main}. MedRouter separates specialist selection from answer generation into two components: an embedding-based Router and an LLM generator. We formulate specialist selection as a multi-label prediction problem and propose \textit{SCALE} (\textit{Specialist Competence-Aware Learning}), a two-stage framework for training the Router. 

\subsection{MedRouter: Specialist Selection and Answer Generation}

\textbf{Specialist selection.}
Let \(q\) denote a medical question, \(\mathcal{S}=\{s_1,\ldots,s_K\}\) a pool of \(K\) frozen specialist LLMs, and \(A\) the ground-truth answer. The embedding encoder \(E_{\psi}\) maps \(q\) to a representation, which is fed into a multi-label MLP Router \(g_{\phi}\) to produce one logit per specialist:
\begin{equation}
\mathbf{z}(q)
=
g_{\phi}\!\left(E_{\psi}(q)\right)
\in \mathbb{R}^{K},
\qquad
p_i(q)=\sigma\!\left(z_i(q)\right).
\end{equation}
Here, \(p_i(q)\) is the predicted call score for specialist \(s_i\). Separate sigmoid outputs allow more than one specialist to be selected for the same question. At inference, the Router selects all specialists whose scores meet or exceed the threshold \(\tau=0.5\):\(\mathcal{S}_{q}=\{s_i\in\mathcal{S}:p_i(q)\geq\tau\}\). If no specialist meets the threshold, it selects the highest-scoring specialist, i.e., \(\mathcal{S}_{q}=\{\arg\max_{s_i\in\mathcal{S}}p_i(q)\}\).

\textbf{Answer generation.}
Each selected specialist produces a response $s_i(q)$. We collect these responses as $I_q=\{s_i(q):s_i\in S_q\}$ and provide them, together with the original question, to the generator: \(\hat{A}_q=G(q,I_q)\). The generator produces the final answer from this input rather than returning a selected specialist's answer directly.

\subsection{Specialist Competence-Aware Learning}
Specialist correctness provides a starting point for learning which models to consult, but it does not fully describe how their responses affect the generator. A specialist may reach an incorrect final answer yet provide information that helps the generator answer correctly. 
\textit{SCALE} therefore first trains the router using individual specialist correctness labels, then refines its selections using generator-level feedback. Throughout training, the embedding encoder, specialist LLMs, and answer generator remain frozen; only the Router parameters \(\phi\) are updated.

\subsubsection{Stage I: Learning Specialist Competence}
Let $\mathcal{D}=\{(q_j,A_j)\}_{j=1}^{N}$ denote the training set, where $A_j$ is the reference answer to question $q_j$. For a model response $r$, let $C(r,A)\in\{0,1\}$ indicate whether its answer is correct with respect to reference $A$. Each specialist receives a binary label $y_{ji}=C(s_i(q_j),A_j)$ for each training question. We train the router with binary cross-entropy on these labels:
\begin{equation}
\mathcal{L}_{\mathrm{FT}}
=
\frac{1}{NK}
\sum_{j=1}^{N}\sum_{i=1}^{K}
\operatorname{BCEWithLogits}(z_{ji},y_{ji}),
\end{equation}
where \(N\) is the number of training questions and \(K\) is the number of specialists.
A question may have several positive labels, allowing the router to learn the competence of multiple specialists rather than choose a single training target. This stage evaluates specialists independently; it does not use the generator's answer to score their responses.


\subsubsection{Stage II: Learning from Generator Outcomes}
\label{sec:scale_stage2}
Starting from the Stage-I router, we use reinforcement learning to evaluate specialist selections through the frozen generator.

\textbf{Routing actions.}
During training, we sample each specialist independently according to its call probability, $a_{qi}\sim\operatorname{Bernoulli}(p_i(q))$. The resulting action $\mathbf{a}_q=(a_{q1},\ldots,a_{qK})$ specifies the selected set $S_q(\mathbf{a}_q)=\{s_i\in S:a_{qi}=1\}$. Let $I_q(\mathbf{a}_q)$ denote the responses returned by these specialists. Sampling allows training to evaluate different selections, whereas inference uses the threshold-based rule described above.

\textbf{Performance Gain Reward.}
For a training question with reference answer $A$, the generator produces answers with and without the selected responses:
\begin{equation}
    \hat{A}_q^{+}=G(q,I_q(\mathbf{a}_q)),
    \qquad
    \hat{A}_q^{-}=G(q,\varnothing).
    \label{eq:scale_generator_answers}
\end{equation}
We define the Performance Gain Reward (PGR) as
\begin{equation}
    R_{\mathrm{PGR}}(q,\mathbf{a}_q)
    =C(\hat{A}_q^{+},A)
    -\frac{1}{2}C(\hat{A}_q^{-},A).
    \label{eq:scale_pgr}
\end{equation}
The reward is $1$ when only the answer with specialist responses is correct, $0.5$ when both answers are correct, $-0.5$ when only the answer without specialist responses is correct, and $0$ when both are incorrect. The selected set is therefore evaluated through the generator's answer, rather than through the correctness of its individual specialists.

\textbf{Policy optimization.}
The independent Bernoulli decisions give the following log-probability for a routing action:
\begin{equation}
    \log\pi_\phi(\mathbf{a}_q\mid q)
    =\sum_{i=1}^{K}\left[
        a_{qi}\log p_i(q)
        +(1-a_{qi})\log(1-p_i(q))
    \right].
    \label{eq:scale_routing_policy}
\end{equation}
We update the router using REINFORCE with a learned value baseline \citep{williams1992simple,sutton1999policy}:
\begin{equation}
    \mathcal{L}_{\mathrm{policy}}
    =-\mathbb{E}_{q,\mathbf{a}_q}\left[
        \left(R_{\mathrm{PGR}}(q,\mathbf{a}_q)-V_\omega(q)\right)
        \log\pi_\phi(\mathbf{a}_q\mid q)
    \right],
    \label{eq:scale_policy_loss}
\end{equation}
where the expectation is over training questions and sampled routing actions. The value head $V_\omega(q)$ estimates the expected reward for question $q$ and serves as a baseline to reduce gradient variance. Subtracting this baseline gives the feedback used to increase or decrease the probability of the sampled selection.

\section{Experiments}
\label{sec:experiments}

\subsection{Experimental Setups}

\noindent\textbf{Router and Generator LLMs.}
MedRouter consists of an embedding-based router and an LLM answer generator. We use the frozen \textit{Qwen3-VL-Embedding-2B} encoder \citep{li2026qwen3} to obtain a 2,048-dimensional last-token representation of each question. The trainable multi-label routing head applies layer normalization, a linear projection to 256 dimensions, a GELU activation, and a final linear layer that outputs one logit for each of the six specialists. For the generator, we evaluate \textit{Qwen3.5-2B} \citep{yang2025qwen3} and \textit{Gemma-4-E2B} \citep{team2026gemma}.

\noindent\textbf{Medical Specialist LLMs.}
Our routing system operates over a pool of medical specialists comprising both text-only LLMs and vision-language models. The text-only specialists are BioMistral-7B \citep{labrak2024biomistral}, MediPhi-3.8B \citep{corbeil2025mediphi}, and OpenBioLLM-8B \citep{aaditya2024openbiollm}; the multimodal specialists are MedGemma-4B \citep{sellergren2025medgemma}, HuatuoGPT-Vision-7B \citep{chen2024huatuogptvision}, and MediX-R1-8B \citep{mullappilly2026medixr1}. Although broadly comparable in parameter scale, these specialists differ in their training corpora, architectures, and training procedures, which may contribute to their distinct performance profiles.

\noindent\textbf{Datasets.}
We evaluate MedRouter and the baselines on four text-based medical QA benchmarks: MedQA \citep{jin2020disease}, MedMCQA \citep{pal2022medmcqa}, MedCaseReasoning \citep{wu2025medcasereasoning}, and the text subset of MedXpertQA \citep{zuo2025medxpertqa}. For multimodal medical QA, we use SLAKE \citep{liu2021slake}, PathVQA \citep{he2021pathvqa}, VQA-RAD \citep{lau2018dataset}, and the multimodal subset of MedXpertQA \citep{zuo2025medxpertqa}. Since MedXpertQA does not provide a training set, we use both its text and multimodal subsets exclusively for out-of-distribution (OOD) evaluation.

\noindent\textbf{Baselines.}
We compare MedRouter against three groups of baselines: non-routing methods, ML-based routing methods, and LLM-based routing.

\begin{itemize}[leftmargin=*, itemsep=0pt, labelsep=5pt, topsep=0pt]

    \item \textbf{Non-routing methods.}
    These methods use a single generator to answer each question without consulting the specialist pool.
    \textbf{Direct} prompts the generator to answer without retrieval or fine-tuning.
    \textbf{RAG} provides the three most similar training questions and their gold answers as in-context examples. We encode questions with \textit{Qwen3-VL-Embedding-2B}, L2-normalize the resulting embeddings, and retrieve training examples based on cosine similarity.
    \textbf{SFT} fine-tunes the generator on the training split, computing loss only on the completion. For multimodal models, images are included during fine-tuning.
    \textbf{RL} trains the generator with GRPO \citep{shao2024deepseekmath} on the training split using a correctness-only reward: \(+1\) for an answer judged correct in the \texttt{<answer>} block, \(0\) for an answer judged incorrect, and \(-1\) if the block is missing or contains the prompt placeholder.
    
    \item \textbf{ML-based routing methods.}
    These baselines select specialists or aggregate their responses using machine-learning-based and voting-based strategies. The ML-based Routers use \textit{Qwen3-VL-Embedding-2B} as the text encoder.
    \textbf{Ensemble} queries every specialist, normalizes their answers by lowercasing them and removing framing text, and selects the answer with the largest group of matching responses.
    \textbf{KNN Router} embeds each question, retrieves its \(k=50\) nearest training examples, and selects the specialist that answered the largest number of retrieved examples correctly.
    \textbf{MLP Router} uses representations from the frozen encoder as input to a feed-forward network, which predicts the probability that each specialist will answer correctly. It selects all specialists whose scores meet the threshold \(\tau=0.5\), defaulting to the highest-scoring specialist if none meets the threshold. Its output head is multi-label, allowing it to select multiple specialists for a question.
    \textbf{Encoder Router} fine-tunes the encoder end-to-end using a multiclass classification head and cross-entropy loss, allowing us to assess the effect of encoder fine-tuning relative to the MLP Router.
    
    \item \textbf{LLM-based routing methods.}
    These methods use an LLM to select specialists, aggregate their responses, or generate the final answer.
    \textbf{LLM Ensemble} provides the question and all specialist responses to an LLM, which generates the final answer. Unlike string voting, it can reconcile paraphrases, discount implausible responses, or select a minority answer, but it queries the entire specialist pool for every question.
    \textbf{Prompt LLM} receives the question and a description of each specialist, selects one specialist, and returns its answer verbatim. It serves as a direct prompting-based LLM routing baseline.
    \textbf{Router-R1} \citep{zhang2025router} is an RL-trained LLM router optimized using only a final-answer reward: \(-1\) for a missing or placeholder answer, \(0\) for an answer judged incorrect, and \(+1\) for an answer judged correct. The LLM actively selects and calls specialists iteratively, incorporates the returned information, and generates the final answer, keeping routing and generation within a single model.
\end{itemize}

\noindent\textbf{Training and Evaluation.}
We use the full test set of each benchmark for evaluation. For training, we use 10,000 examples each from MedQA, MedMCQA, and MedCaseReasoning, and 4,918, 19,654, and 1,793 examples from SLAKE, PathVQA, and VQA-RAD, respectively. This yields 30,000 text-based and 26,365 multimodal training examples, which we combine during training. We randomly sample 200 examples from each dataset's training split to form a 1,200-example development set, corresponding to approximately 2.1\% of the selected training examples. The combined test sets contain 8,803 text-only and 10,231 image-based questions, for a total of 19,034 questions. We use \textit{GPT-5.6-luna} as an LLM judge to assess whether each generated answer is semantically consistent with the ground-truth answer in its medical context.
Further details on training, evaluation, and the \textit{MedRouter} and baseline configurations are provided in Appendix~\ref{sec:exp_details}.

\subsection{Main Results}

\begin{table*}[t]
\centering
\small
\caption{Accuracy (\%) on text-based and multimodal medical QA benchmarks. Within each LLM generator block, the best result for each dataset and the average is shown in \textbf{bold}. The relative improvement of \textit{MedRouter} over the second-highest average in the same block is shown in \gain{green}.}
\label{tab:main_results}
\resizebox{\textwidth}{!}{%
\begin{tabular}{lccccccccc}
\toprule
& \multicolumn{4}{c}{\textbf{Text-Based Medical QA}}
& \multicolumn{4}{c}{\textbf{Multimodal Medical QA}}
& \\
\cmidrule(lr){2-5} \cmidrule(lr){6-9}
& \textbf{MedQA} & \textbf{MedMCQA} & \textbf{MedCaseRea.} & \textbf{MedX.-Text}
& \textbf{SLAKE} & \textbf{PathVQA} & \textbf{VQA-RAD} & \textbf{MedX.-MM} & \textbf{Average} \\
\midrule
\rowcolor{gray!30}
\quad Best-in-Pool & 65.2 & 59.3 & 15.1 & 15.4 & 67.0 & 41.7 & 57.9 & 27.4 & 43.6 \\
\midrule
\multicolumn{10}{l}{\textbf{\textit{ML Routing Methods}}} \\
\quad Ensemble       & 61.6 & 59.0 & 15.7 & 14.5 & 66.2 & 39.5 & 50.8 & 27.0 & 41.8 \\
\quad KNN Router     & 61.0 & 56.5 & 12.3 & 16.0 & 70.0 & 41.3 & 56.7 & 26.1 & 42.5 \\
\quad MLP Router     & 57.7 & 52.2 & 10.7 & 13.7 & 70.7 & 40.9 & 56.1 & 24.9 & 40.9 \\
\quad Encoder Router & 62.5 & 55.9 & 14.7 & 14.9 & 71.6 & 40.0 & 54.3 & 27.3 & \textcolor{langgreen}{42.6} \\
\midrule
\multicolumn{10}{l}{\textbf{\raisebox{-0.2\height}{\includegraphics[width=1.3em]{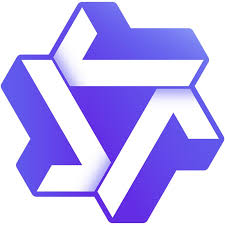}}\ Qwen3.5-2B}} \\
\multicolumn{10}{l}{\textbf{\textit{Non-Routing Methods}}} \\
\quad Direct & 52.0 & 48.8 & 7.4 & 14.6 & 59.2 & 37.3 & 51.6 & 26.1 & 37.1 \\
\quad RAG    & 51.5 & 43.7 & 11.4 & 15.5 & 70.7 & 37.8 & 44.1 & 25.9 & 37.5 \\
\quad SFT    & 51.1 & 47.8 & 9.1 & 16.9 & \textbf{74.0} & \textbf{44.9} & 45.4 & 25.6 & 39.4 \\
\quad RL     & 51.7 & 48.9 & 10.1 & 14.4 & 60.2 & 42.3 & 53.9 & 26.0 & 38.4 \\
\midrule
\multicolumn{10}{l}{\textbf{\textit{Routing Baselines}}} \\
\quad Ensemble   & 54.7 & 50.4 & 14.2 & 14.7 & 59.2 & 37.5 & 53.2 & 25.9 & 38.7 \\
\quad Prompt LLM & 53.2 & 49.0 & 10.4 & 13.1 & 65.6 & 39.1 & \textbf{56.0} & 24.5 & 38.9 \\
\quad Router-R1  & 50.8 & 46.1 & 12.4 & 16.0 & 52.8 & 34.9 & 46.0 & 20.3 & 34.9 \\
\midrule
\rowcolor{langlightgreen!80}
\quad \textbf{MedRouter} & \textbf{71.8} & \textbf{59.9} & \textbf{19.6} & \textbf{17.3} & 72.0 & 43.4 & 55.1 & \textbf{27.9} & \textbf{45.9}\,\gain{(+7.7\%)} \\
\midrule
\multicolumn{10}{l}{\textbf{\raisebox{-0.2\height}{\includegraphics[width=1.3em]{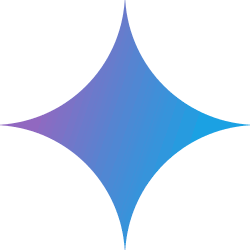}}\ Gemma-4-E2B}} \\
\multicolumn{10}{l}{\textbf{\textit{Non-Routing Methods}}} \\
\quad Direct & 63.0 & 49.3 & 10.7 & 14.7 & 50.6 & 34.7 & 48.1 & 24.1 & 36.9 \\
\quad RAG    & 58.1 & 47.7 & 12.8 & 15.4 & 53.6 & 39.7 & 55.0 & 25.1 & 38.4 \\
\quad SFT    & 50.4 & 45.4 & 8.8 & 14.2 & 53.3 & 40.1 & 51.9 & 23.0 & 35.9 \\
\quad RL     & 53.3 & 44.2 & 9.4 & 13.3 & 48.1 & 37.3 & 52.7 & 20.3 & 34.8 \\
\midrule
\multicolumn{10}{l}{\textbf{\textit{Routing Baselines}}} \\
\quad LLM Ensemble & 65.1 & 57.4 & 16.7 & 15.2 & 51.5 & 36.4 & \textbf{55.4} & 24.9 & 40.3 \\
\quad Prompt LLM   & 54.7 & 50.1 & 11.2 & 12.9 & 48.3 & \textbf{45.3} & 52.6 & 26.6 & 37.7 \\
\quad Router-R1    & 52.6 & 46.7 & 12.3 & 14.5 & 46.5 & 42.9 & 40.3 & 25.8 & 35.2 \\
\midrule
\rowcolor{langlightgreen!80}
\quad \textbf{MedRouter} & \textbf{74.3} & \textbf{63.0} & \textbf{20.3} & \textbf{16.5} & \textbf{67.9} & 44.1 & 53.9 & \textbf{29.1} & \textbf{46.1}\,\gain{(+8.2\%)} \\
\bottomrule
\end{tabular}%
}
\end{table*}

Table~\ref{tab:main_results} reports accuracy across eight medical QA benchmarks. MedRouter achieves an average accuracy of 45.9\% with \textit{Qwen3.5-2B} and 46.1\% with \textit{Gemma-4-E2B}, representing relative improvements of 7.7\% and 8.2\%, respectively, over the strongest routing baseline, Encoder Router (42.6\%). Both variants also exceed the 43.6\% average of Best-in-Pool, which uses the strongest individual specialist for each dataset. The gains are especially pronounced on text-based QA and extend to both out-of-distribution MedXpertQA subsets. With \textit{Qwen3.5-2B}, MedRouter improves accuracy on MedQA, which draws on medical licensing exams to assess broad medical knowledge, from the best routing baseline's 62.5\% to 71.8\%, a relative gain of 14.9\%. On MedCaseReasoning, which evaluates diagnosis from clinical case reports often involving rare or complex conditions and long-tail reasoning challenges, it achieves a relative accuracy gain of 24.8\%. These results demonstrate that \textit{MedRouter} can effectively improve medical QA performance by selecting and integrating specialist responses at the question level. Beyond these performance gains, our experiments also reveal several valuable insights.

\noindent\textbf{Leveraging Knowledge Encoded in Specialist LLMs.}
The limited gains from directly post-training the generator on our available QA training data suggest that this approach alone is insufficient in our setting. Many current LLMs have already been trained on large amounts of publicly available web data. Meanwhile, existing specialist LLMs may encode medical knowledge acquired from sources not represented in our QA training data. This motivates us to draw on knowledge encoded in existing LLMs rather than relying solely on public datasets. For example, \textit{MedGemma} was trained in part on licensed and de-identified medical datasets \citep{sellergren2025medgemma}. Rather than attempting to transfer all such knowledge into a single generator, \textit{MedRouter} accesses it through specialist responses to improve performance on medical reasoning tasks.

\noindent\textbf{Decoupling Routing and Generation.}
Our results suggest that jointly learning specialist selection and answer generation within a single LLM can be challenging. For example, Router-R1 \citep{zhang2025router} uses an RL-trained LLM router with a final-answer reward to select specialists and reason over their responses iteratively. In our experiments, this training is unstable: the final-answer reward provides only an indirect and sparse signal for routing, and jointly optimizing selection and reasoning may make it difficult for the model to learn which specialist to consult for each question. The routing policy can consequently collapse toward a narrow set of specialists, degrading performance. Table~\ref{tab:main_results} also shows that the ML-based routing methods generally outperform the LLM-based routing baselines in our evaluation. Therefore, in designing \textit{MedRouter}, we use an embedding-based router that can be trained directly to predict specialist competence and optimize selections for downstream answer accuracy. A separate LLM generator integrates the selected responses to address diverse questions without also having to make routing decisions. By decoupling routing from generation, we allow each component to focus on a distinct role and achieve better performance.

\textit{MedRouter} does not achieve the best result on every multimodal benchmark. Specialist responses may omit relevant visual evidence or introduce conflicting information, and our pool contains only three vision-capable specialists. Moreover, because the generator remains frozen, training the router does not directly improve the generator’s visual understanding. These factors may limit its ability to answer image-based questions correctly.

\subsection{Ablation Study of \textit{SCALE}}

\begin{table*}[t]
\centering
\small
\caption{Ablation study of \textit{SCALE} with \textit{Qwen3.5-2B} as the generator. The best result in each column is shown in \textbf{bold}. The relative decrease in average accuracy from the full \textit{SCALE} is shown in \loss{red}.}
\label{tab:scale_ablation}
\resizebox{\textwidth}{!}{%
\begin{tabular}{lccccccccc}
\toprule
& \multicolumn{4}{c}{\textbf{Text-Based Medical QA}}
& \multicolumn{4}{c}{\textbf{Multimodal Medical QA}}
& \\
\cmidrule(lr){2-5} \cmidrule(lr){6-9}
& \textbf{MedQA} & \textbf{MedMCQA} & \textbf{MedCaseRea.} & \textbf{MedX.-Text}
& \textbf{SLAKE} & \textbf{PathVQA} & \textbf{VQA-RAD} & \textbf{MedX.-MM} & \textbf{Average} \\
\midrule
\rowcolor{langlightgreen!80}
\quad \textbf{MedRouter w/ SCALE} & \textbf{71.8} & 59.9 & \textbf{19.6} & \textbf{17.3} & \textbf{72.0} & \textbf{43.4} & \textbf{55.1} & \textbf{27.9} & \textbf{45.9} \\
\midrule
\quad w/ Stage I FT only & 69.3 & 57.2 & 17.6 & 16.5 & 69.3 & 41.1 & 51.4 & 26.0 & 43.6\,\loss{(-5.0\%)} \\
\quad w/ Stage II RL only & 70.1 & 59.1 & 17.7 & 16.7 & 71.2 & 41.3 & 54.9 & 27.2 & 44.8\,\loss{(-2.4\%)} \\
\quad w/ Stage II RL using correctness reward & 70.6 & \textbf{60.0} & 18.4 & 16.6 & 69.7 & 42.6 & 52.5 & 27.0 & 44.7\,\loss{(-2.6\%)} \\
\bottomrule
\end{tabular}%
}
\end{table*}

To assess the effectiveness of \textit{SCALE}, we ablate each training stage and the Performance Gain Reward (PGR), as shown in Table~\ref{tab:scale_ablation}. The full two-stage framework achieves the highest average accuracy of 45.9\% and the best result on seven of the eight benchmarks.

Using Stage I fine-tuning alone reduces average accuracy to 43.6\%, while using Stage II reinforcement learning alone yields 44.8\%. Together, these results suggest that correctness supervision provides a useful initialization, while subsequent RL further improves specialist selection for the downstream generator, making the two stages complementary. The gain over Stage I alone is consistent with the motivation for Stage II: whether a specialist answers correctly is an imperfect measure of whether its response contains useful information, as the case study in Section~\ref{sec:case_study} illustrates. Conversely, the lower performance of RL from scratch suggests that learning from final-answer feedback alone is more difficult. In our training runs, we find this setting is also more prone to concentrating calls on a narrow set of specialists, which may limit routing diversity and performance.

Replacing PGR with a correctness-only reward (\(+1\) for a correct final answer and \(0\) otherwise) reduces the average accuracy from 45.9\% to 44.7\%. Unlike the correctness-only reward, PGR uses the generator's answer without specialist information as a question-specific reference, distinguishing cases where specialist information corrects an otherwise incorrect answer from those where the generator can already answer correctly on its own. By explicitly rewarding the improvement attributable to specialist information, PGR provides a more informative learning signal for routing and may reduce variance in RL updates, leading to more stable training and higher final accuracy.

\subsection{Analysis of Routing}

\begin{figure*}[t!]
    \centering
    \includegraphics[width=\textwidth]{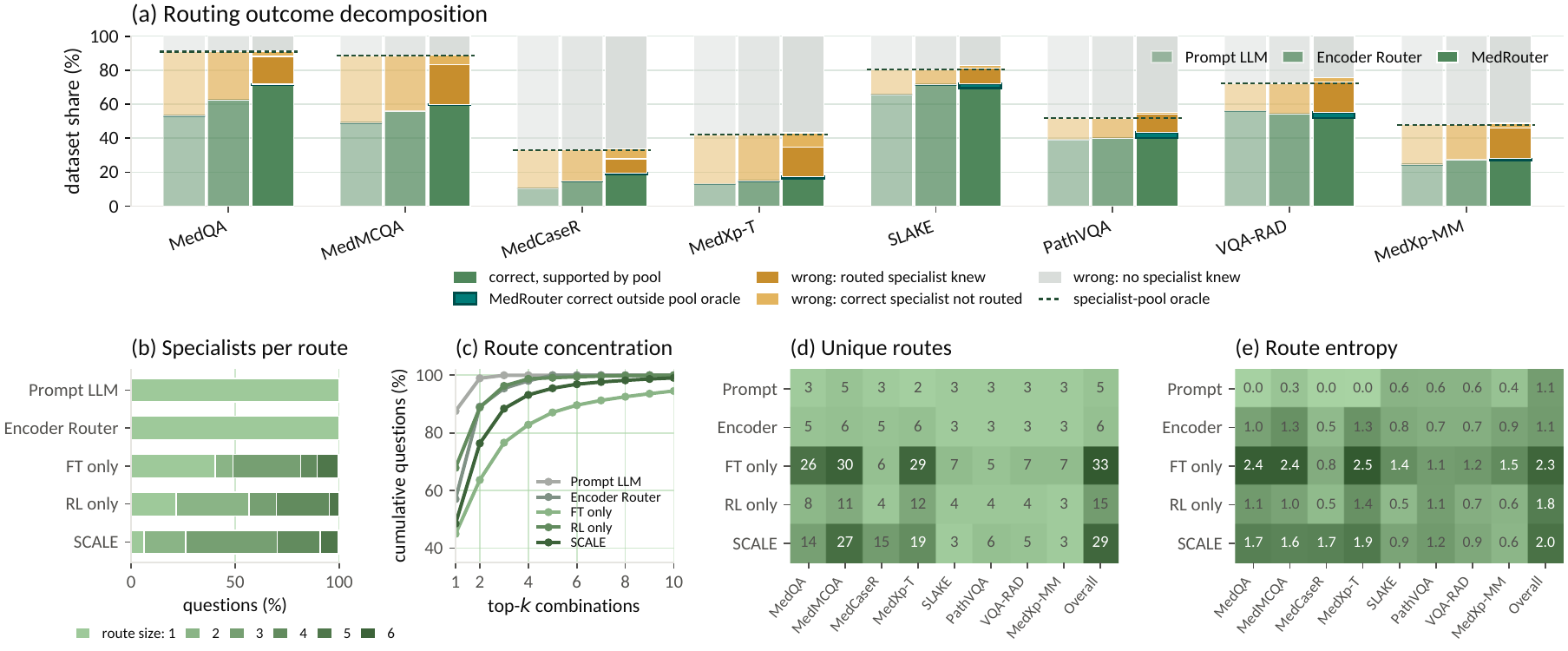}
    \caption{Routing outcomes and specialist selection patterns across eight benchmarks. (a) Decomposition of outcomes for Prompt-LLM, Encoder Router, and \textit{MedRouter}. The dashed line marks the fraction of questions for which at least one specialist knows the answer. (b) Distribution of the number of specialists selected per question. (c) Cumulative fraction of questions assigned to the $k$ most frequent specialist combinations. (d) Number of unique routes on each benchmark and overall. (e) Entropy of the route distribution on each benchmark and overall.}
    \label{fig:routing_analysis}
\end{figure*}

Figure~\ref{fig:routing_analysis}(a) decomposes the outcomes of different routing strategies according to whether the specialist pool contains a knowledgeable specialist and whether the routed information leads to a correct answer. \textit{MedRouter} achieves a larger proportion of correct answers supported by the specialist pool. Notably, \textit{MedRouter} also answers 332 questions correctly even when no specialist in the pool produces the correct answer. This suggests that the generator can benefit from complementary information in imperfect specialist responses and synthesize a correct answer beyond the capability of any individual specialist; we provide an illustrative example in the case study in Appendix~\ref{sec:case_study}. Nevertheless, effectively integrating specialist information remains a bottleneck: some questions are still answered incorrectly despite routing to a knowledgeable specialist, as the LLM generator is frozen and is not optimized within our framework.

Figures~\ref{fig:routing_analysis}(b)--(e) further examine specialist selection and route diversity. \textit{SCALE} often selects multiple specialists and uses 29 distinct routes across the benchmarks. Although the fine-tuning-only variant uses more distinct routes and exhibits higher route entropy than \textit{SCALE}, \textit{SCALE} still maintains substantially greater route diversity and higher overall entropy than the RL-only variant. These results indicate that the combined training procedure preserves diverse specialist combinations while avoiding the substantial reduction in routing diversity observed with RL alone.

\section{Conclusion}
In this paper, we examined how medical specialist LLMs differ in question-answering performance and found that their strengths are complementary at the question level. Motivated by this finding, we introduced \textit{MedRouter}, which separates specialist selection from answer generation, and \textit{SCALE}, a two-stage framework that trains the router using specialist correctness supervision followed by reinforcement learning based on the downstream value of specialist responses. Across eight text-based and multimodal medical QA benchmarks, MedRouter improves average accuracy over the strongest routing baselines while keeping the specialists and generator frozen. These results highlight the potential of question-level routing to make better use of existing medical LLMs and motivate further work on combining their capabilities for more complex medical reasoning tasks.






\clearpage

\bibliography{iclr2027_conference}
\bibliographystyle{iclr2027_conference}

\clearpage

\appendix

\DoToC

\clearpage

\section{AI Use Statement}
\label{sec:ai_use}

In this work, we used generative AI tools for interpreting experimental results and assisting with the writing of the manuscript. We have not used generative AI tools for generating datasets, conducting experiments without author verification, or making final research decisions, and other required disclosure tasks are not applicable to this work. Additionally, we used generative AI tools for brainstorming, literature search, improving readability, polishing text, organizing paper structure, formatting tables, and formatting references. We have reviewed all AI-assisted work. We manually checked technical claims, citations, experimental results, and manuscript content. We take responsibility for the final content of this work, including text, claims, and artifacts produced with the aid of generative AI.

\section{Ethics Statement}
\label{sec:ethics}

This work studies how to route medical questions to specialist LLMs and combine their responses to support medical reasoning. Our experiments use public medical question-answering benchmarks and model-generated responses for research evaluation. These systems are not intended to provide clinical diagnoses, treatment recommendations, or other professional medical advice, and their outputs may be inaccurate, incomplete, or inconsistent. MedRouter should therefore be viewed as a research framework rather than a clinical decision-making tool. Any real-world clinical use would require validation by qualified medical professionals, appropriate privacy protections, and rigorous assessment of safety and potential harm.

\section{Limitations}
\label{sec:limitations}

This work focuses on routing medical questions to specialist LLMs and leveraging their responses to improve answer correctness, but several limitations remain. MedRouter depends on the availability and reliability of its specialist models, and its performance may vary across datasets and deployment settings. Moreover, our evaluation is limited to medical question answering, whereas many real-world medical tasks involve more complex forms of reasoning and information synthesis. Such tasks may require gathering, organizing, and integrating specialist input over multiple steps. Future work could extend MedRouter to these settings and investigate more effective ways to exploit specialist information for complex medical reasoning.

\section{Experimental Setup Details}
\label{sec:exp_details}

\noindent\textbf{Specialist Response Generation.}
Each specialist answers every question once offline, and we cache its response for use during training and evaluation. This ensures that all routing methods receive the same specialist information. We generate responses with temperature $0.0$, a maximum of $1536$ tokens, and thinking mode disabled. The prompt requires the format \verb|<reasoning>...</reasoning><answer>...</answer>|. When a specialist is selected, the generator receives only its reasoning, not its final answer.

\noindent\textbf{\textit{MedRouter} Training.}
In Stage I of fine-tuning, we train the policy head with binary cross-entropy to predict whether each specialist answers a question correctly. We use AdamW with a learning rate of $2 \times 10^{-5}$ and a batch size of $16$. Training runs for up to $3$ epochs, with evaluation every $100$ steps and early stopping after $3$ evaluations without improvement. The resulting checkpoint initializes Stage II and also serves as a standalone baseline. In Stage II of RL, we randomly initialize the value head. We use AdamW with learning rates of $4 \times 10^{-5}$ for the policy and $1 \times 10^{-4}$ for the value head, a batch size of $128$ questions, an entropy coefficient of $\beta=0.01$, and gradient clipping at a norm of $1.0$. Batches reflect the proportions of the training sources. Training runs for up to $10$ epochs, with development-set evaluation and checkpointing every $100$ steps.

\noindent\textbf{Baselines.}
We compare \textit{MedRouter} with several routing baselines. The KNN router retrieves the $k=50$ nearest training questions using frozen \textit{Qwen3-VL-Embedding-2B} embeddings, while the MLP router uses the same architecture as the policy head in \textit{MedRouter} over the same embedding features. RAG use original question to retrieve the top $k=3$ training questions using the same embeddings and provides them as context for generation. For Router-R1, we remove the cost reward and optimize the router using only the answer-correctness reward. The SFT and RL  baselines use the default training configurations provided by VERL \citep{sheng2024hybridflow}.
    
\noindent\textbf{Infrastructure.}
All training and evaluation runs use a single NVIDIA GH200 GPU. Specialists and the generator are served with vLLM \citep{kwon2023efficient}.

\section{Performance of Medical Specialist LLMs}
\label{sec:specialists}

Table~\ref{tab:specialist_results} reports the performance of each model across the eight medical QA benchmarks. Among the medical specialists, MediX-R1-8B performs best on all four text-based datasets, achieving an average accuracy of 38.7\%. On multimodal QA, the strongest specialist varies by dataset: MediX-R1-8B leads on SLAKE and MedXpertQA-MM, while MedGemma-4B leads on PathVQA and VQA-RAD and achieves the highest multimodal average of 46.8\%. Selecting the best specialist separately for each multimodal dataset raises the average to 48.5\%, compared with 46.8\% for the best single specialist. For text-based QA, the dataset-level Best-in-Pool average equals MediX-R1-8B's average; the potential benefit of routing within these datasets instead arises from the question-level differences analyzed in Figure~\ref{fig:analysis}. Text-only specialists are not evaluated on multimodal datasets because they do not accept image inputs

\begin{table*}[t]
\centering
\small
\caption{Accuracy (\%) of individual models on text-based and multimodal medical QA benchmarks. Best-in-Pool selects the highest-scoring medical specialist separately for each dataset. Bold values indicate the best result among individual medical specialists in each column.}
\label{tab:specialist_results}
\resizebox{\textwidth}{!}{%
\begin{tabular}{lcccccccccc}
\toprule
& \multicolumn{5}{c}{\textbf{Text-Based Medical QA}}
& \multicolumn{5}{c}{\textbf{Multimodal Medical QA}} \\
\cmidrule(lr){2-6} \cmidrule(lr){7-11}
& \textbf{MedQA} & \textbf{MedMCQA} & \textbf{MedCaseRea.} & \textbf{MedX.-Text} & \textbf{Average}
& \textbf{SLAKE} & \textbf{PathVQA} & \textbf{VQA-RAD} & \textbf{MedX.-MM} & \textbf{Average} \\
\midrule
\rowcolor{gray!30}
\quad Best-in-Pool & 65.2 & 59.3 & 15.1 & 15.4 & 38.7 & 67.0 & 41.7 & 57.9 & 27.4 & 48.5 \\
\midrule
\multicolumn{11}{l}{\textbf{\textit{General-Purpose LLMs}}} \\
\quad \raisebox{-0.2\height}{\includegraphics[width=1.3em]{figures/icons/qwen.png}}\ Qwen3.5-2B & 52.0 & 48.8 & 7.4 & 14.6 & 30.7 & 59.2 & 37.3 & 51.6 & 26.1 & 43.6 \\
\quad \raisebox{-0.2\height}{\includegraphics[width=1.3em]{figures/icons/gemini.png}}\ Gemma-4-E2B & 63.0 & 49.3 & 10.7 & 14.7 & 34.4 & 50.6 & 34.7 & 48.1 & 24.1 & 39.4 \\
\midrule
\multicolumn{11}{l}{\textbf{\textit{Medical Specialist LLMs}}} \\
\quad BioMistral-7B  & 35.9 & 36.0 & 10.0 & 12.2 & 23.5 & -- & -- & -- & -- & -- \\
\quad MediPhi-3.8B   & 53.1 & 50.2 & 10.3 & 12.7 & 31.6 & -- & -- & -- & -- & -- \\
\quad OpenBioLLM-8B  & 52.6 & 49.9 & 9.6 & 9.3 & 30.3 & -- & -- & -- & -- & -- \\
\quad MedGemma-4B   & 56.0 & 47.7 & 9.6 & 14.5 & 31.9 & 63.6 & \textbf{41.7} & \textbf{57.9} & 23.9 & \textbf{46.8} \\
\quad HuatuoGPT-V-7B & 56.6 & 52.2 & 9.5 & 12.4 & 32.7 & 51.8 & 34.9 & 42.6 & 22.6 & 38.0 \\
\quad MediX-R1-8B    & \textbf{65.2} & \textbf{59.3} & \textbf{15.1} & \textbf{15.4} & \textbf{38.7} & \textbf{67.0} & 40.9 & 49.9 & \textbf{27.4} & 46.3 \\
\bottomrule
\end{tabular}%
}
\end{table*}

\clearpage

\section{Case Study: Integrating Complementary Specialist Information}
\label{sec:case_study}

We examine a MedXpertQA-Text question for which the baseline \texttt{Qwen3.5-2B} generator answers incorrectly in isolation but reaches the correct answer when provided with routed specialist reasoning traces. Notably, all routed specialists select incorrect final options, yet their intermediate reasoning contains complementary diagnostic cues that the generator can integrate to recover the correct solution. The quoted outputs retain their original wording; \texttt{[...]} denotes omitted passages.

\begin{tcolorbox}[enhanced,breakable,colback=gray!5,colframe=gray!20,coltitle=black,fonttitle=\bfseries,title={Question: MedXpertQA-Text \#1464}]
A 56-year-old female recent immigrant from rural China is evaluated after her husband brings her to the physician with a two-day history of confusion, fever, headaches, and malaise. Her vaccination history is unknown. She presents with a temperature of 39.1~$^\circ$C (102.4~$^\circ$F), nuchal rigidity, and photophobia. CSF analysis reveals a neutrophil count of 1,500/mm$^3$. While initial cultures on \textbf{Thayer--Martin and sheep blood agar show no growth}, subsequent heating of the sheep blood agar yields multiple non-hemolytic, opaque, cream-colored colonies. What is the most likely morphological and staining characteristic of the causative organism?

\medskip
\noindent
(A) Gram-negative coccobacillus \quad
(B) Gram-positive, lancet-shaped diplococcus \quad
(C) Gram-variable, filamentous organism \quad
(D) Gram-negative, facultative intracellular bacillus \quad
(E) Gram-negative diplococcus \quad
(F) Gram-positive, pleomorphic rod \quad
(G) Gram-positive, spore-forming rod \quad
(H) Gram-negative, encapsulated bacillus \quad
(I) Gram-positive, branching bacillus \quad
(J) Gram-negative, aerobic coccus

\medskip
\noindent
\textbf{Ground truth: (A) Gram-negative coccobacillus.}
\end{tcolorbox}

The culture phenotype points to \textit{Haemophilus influenzae}, a fastidious Gram-negative coccobacillus that requires X (hemin) and V (NAD) factors for growth. Heating blood during the preparation of chocolate agar lyses red blood cells, making these growth factors available, while also inactivating heat-labile NADases that degrade factor V. The absence of growth on selective Thayer--Martin agar further weighs against \textit{Neisseria meningitidis}.

Without specialist information, the baseline \texttt{Qwen3.5-2B} generator incorrectly selects \textbf{(B) Gram-positive, lancet-shaped diplococcus}, consistent with \textit{Streptococcus pneumoniae} rather than \textit{H. influenzae}.

All three routed specialists below produce incorrect final option selections. Their intermediate reasoning traces, however, contain distinct diagnostic signals.

\begin{tcolorbox}[enhanced,breakable,colback=yellow!4,colframe=yellow!25,coltitle=black,fonttitle=\bfseries,title={Specialist Outputs}]

\textbf{OpenBioLLM-8B.}

\textbf{Raw Output:}
\begin{lstlisting}[basicstyle=\ttfamily\footnotesize,columns=fullflexible,breaklines=true,breakatwhitespace=false]
The most likely organism in this scenario is Haemophilus influenzae, which is known to grow better on enriched media such as Fildes' agar or chocolate agar, and can be identified by its characteristic 'drumstick' appearance on Gram stain.
[...]
Therefore, the most likely morphological and staining characteristic of the causative organism is a Gram-negative, aerobic coccus.
\end{lstlisting}

\textbf{Analysis.} OpenBioLLM-8B correctly identifies the causative organism as \textit{H. influenzae}, but subsequently mischaracterizes its morphology. \textit{H. influenzae} is a Gram-negative coccobacillus rather than a coccus, and the ``drumstick'' appearance is classically associated with the terminal spores of \textit{Clostridium tetani}. Thus, the useful signal in this trace is the organism-level identification, whereas its morphological characterization is misleading.

\medskip
\textbf{MedGemma-4B.}

\textbf{Raw Output:}
\begin{lstlisting}[basicstyle=\ttfamily\footnotesize,columns=fullflexible,breaklines=true,breakatwhitespace=false]
This is highly suggestive of *Legionella* species.
[...]
(E) Gram-negative diplococcus: This is characteristic of *Neisseria gonorrhoeae* or *Neisseria meningitidis*, which would typically show growth on Thayer-Martin agar.
[...]
(H) Gram-negative, encapsulated bacillus: This is characteristic of *Haemophilus influenzae* or *Klebsiella pneumoniae*, which would typically show growth on blood agar.
[...]
Therefore, the most likely answer is (D).
\end{lstlisting}

\textbf{Analysis.} MedGemma-4B incorrectly concludes (D), but its reasoning provides useful elimination evidence: it notes that \textit{Neisseria} species would typically grow on Thayer--Martin agar, whereas no such growth is observed in the vignette. It also considers \textit{H. influenzae} in its differential diagnosis. Although its claim that \textit{H. influenzae} would typically grow on unheated blood agar is incorrect, the trace nevertheless contributes a useful cue for distinguishing \textit{H. influenzae} from \textit{Neisseria}.

\medskip
\textbf{MediX-R1-8B.}

\textbf{Raw Output:}
\begin{lstlisting}[basicstyle=\ttfamily\footnotesize,columns=fullflexible,breaklines=true,breakatwhitespace=false]
[...] it allows growth of organisms like Neisseria that are inhibited at room temperature or require specific conditions.
[...]
Therefore, the most likely morphological and staining characteristic is Gram-negative diplococcus.
\end{lstlisting}

\textbf{Analysis.} MediX-R1-8B incorrectly interprets the growth after heating as evidence for \textit{Neisseria} and consequently selects option (E). Unlike the preceding traces, this reasoning provides a misleading diagnostic signal that the generator must reject.

\end{tcolorbox}

The specialist traces therefore contain a mixture of useful and misleading evidence. OpenBioLLM-8B supplies the correct organism-level hypothesis (\textit{H. influenzae}) but an incorrect morphology, whereas MedGemma-4B supplies useful negative evidence against \textit{Neisseria} despite reaching an incorrect final answer. MediX-R1-8B instead reinforces an incorrect competing diagnosis. Recovering the correct answer requires selectively retaining the useful intermediate cues, rejecting the specialists' erroneous conclusions, and mapping \textit{H. influenzae} to its correct coccobacillary morphology.

\begin{tcolorbox}[enhanced,breakable,colback=green!3,colframe=green!20,coltitle=black,fonttitle=\bfseries,title={Generator With Routed Specialist Information}]

\textbf{Raw Generator Output:}
\begin{lstlisting}[basicstyle=\ttfamily\footnotesize,columns=fullflexible,breaklines=true,breakatwhitespace=false]
Heating denatures inhibitors (like hemolysin) that prevent growth at room temperature, allowing the organism to thrive.
[...]
While *Neisseria meningitidis* is a strong candidate for meningitis in this demographic, the specific culture behavior (growth only after heating) and the colony description point more strongly toward *Haemophilus influenzae*.
[...]
Therefore, the organism is *Haemophilus influenzae*, which is a Gram-negative coccobacillus.
\end{lstlisting}

\textbf{Answer: (A) Gram-negative coccobacillus.}

\textbf{Analysis.} Given the routed specialist traces, the generator correctly favors \textit{H. influenzae} over \textit{Neisseria} and maps the organism to its correct morphology. In doing so, it retains the useful organism-level hypothesis while rejecting the incorrect morphological characterization provided by OpenBioLLM-8B and the competing \textit{Neisseria} interpretation proposed by MediX-R1-8B. The final answer is therefore correct even though the generator's mechanistic explanation is not entirely precise: while heating does inactivate heat-labile inhibitors, the specific reference to ``hemolysin'' is unsupported. More precisely, heating lyses red blood cells to make X (hemin) and V (NAD) factors available and inactivates heat-labile NADases that would otherwise degrade factor V.
\end{tcolorbox}

This case illustrates a key advantage of integrating specialist reasoning rather than relying solely on specialist final answers: even when every specialist selects an incorrect option, their intermediate traces can contain complementary diagnostic evidence. MedRouter enables the generator to exploit these partial signals while filtering contradictory or erroneous information, yielding a correct answer that neither the baseline generator nor any individual routed specialist produces independently.

\clearpage

\section{Prompts Used in This Work}
\label{sec:prompt}

\begin{figure*}[h]
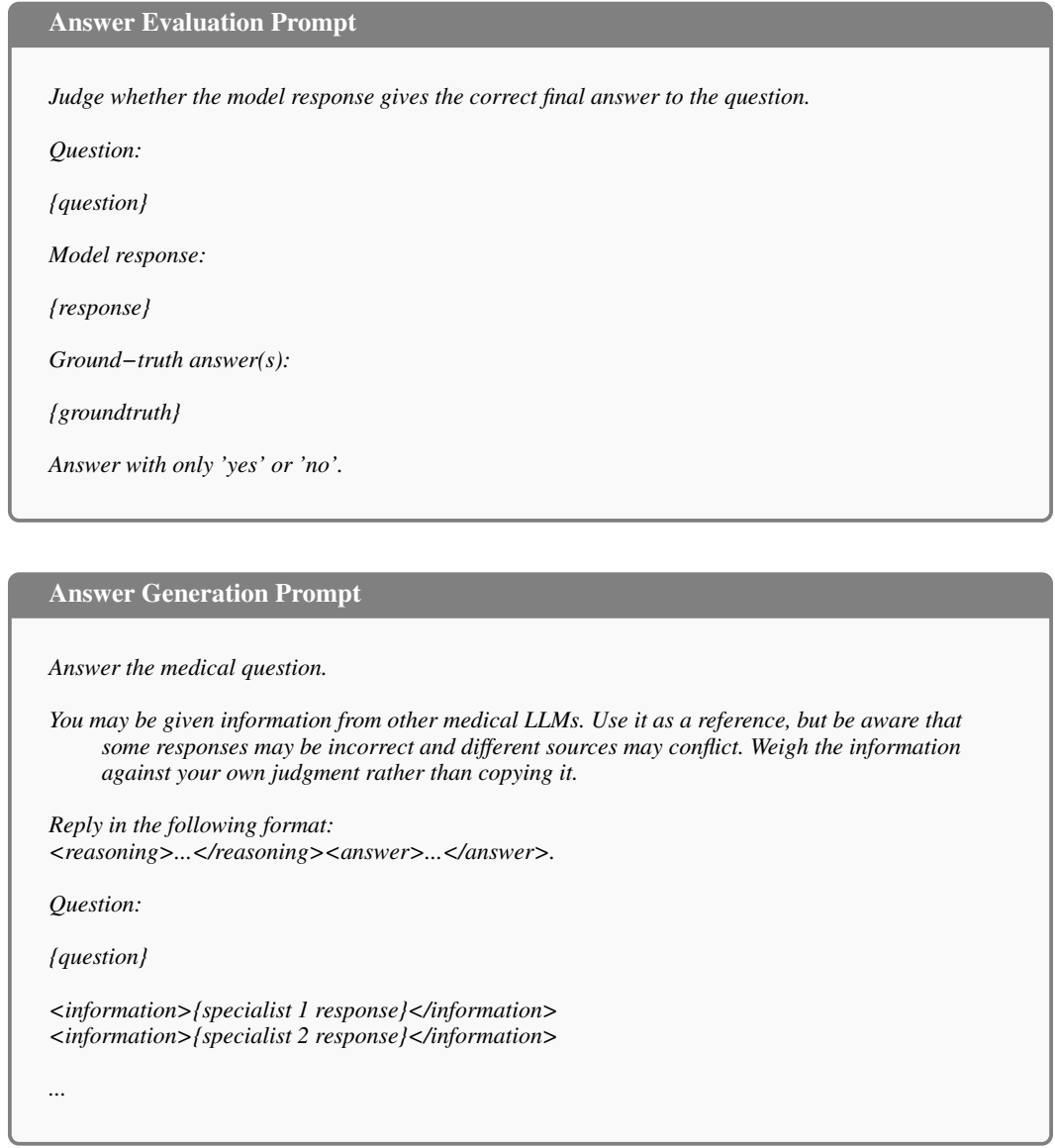

\begin{promptbox}{Answer Evaluation Prompt}
Judge whether the model response gives the correct final answer to the question.

Question:

{question}

Model response:

{response}

Ground-truth answer(s):

{groundtruth}

Answer with only 'yes' or 'no'.
\end{promptbox}

\vspace{0.8em}

\begin{promptbox}{Answer Generation Prompt}
Answer the medical question.

You may be given information from other medical LLMs. Use it as a reference, but be aware that some responses may be incorrect and different sources may conflict. Weigh the information against your own judgment rather than copying it.

Reply in the following format:
<reasoning>...</reasoning><answer>...</answer>.

Question:

{question}

<information>{specialist 1 response}</information>
<information>{specialist 2 response}</information>

...
\end{promptbox}
\caption{Prompts used for answer evaluation and answer generation. Each selected specialist's response is placed in a separate information block.}
\label{fig:prompts}
\end{figure*}

Figure~\ref{fig:prompts} presents the prompts used for answer evaluation and generation. The evaluation prompt asks the judge to compare a model response with the ground-truth answer and return a binary judgment. The generation prompt provides the medical question and, when specialists are selected, their responses in separate \texttt{<information>} blocks. It instructs the generator to consider these responses as potentially useful but fallible evidence, since specialists may be incorrect or disagree.

\end{document}